\documentclass{article}

\usepackage{arxiv}

\usepackage[utf8]{inputenc} 
\usepackage[T1]{fontenc}    
\usepackage{hyperref}       
\usepackage{url}            
\usepackage{booktabs}       
\usepackage{amsfonts}       
\usepackage{nicefrac}       
\usepackage{microtype}      
\usepackage{lipsum}		
\usepackage{graphicx}
\usepackage{doi}
\usepackage{microtype}
\usepackage{graphicx}
\usepackage{subfigure}
\usepackage{booktabs}
\usepackage{amsmath}
\DeclareMathOperator*{\argmax}{arg\,max}
\usepackage{amsfonts} 
\usepackage{algorithm}
\usepackage{algcompatible}
\usepackage{float}

\title{Protein Sequence Design with Batch Bayesian Optimisation}

\author{ {\hspace{1mm}Chuanjiao Zong} \\
	\\
	 \\
}

\renewcommand{\headeright}{preprint under review}
\renewcommand{\undertitle}{preprint under review}

\hypersetup{
pdftitle={A template for the arxiv style},
pdfsubject={q-bio.NC, q-bio.QM},
pdfauthor={David S.~Hippocampus, Elias D.~Striatum},
pdfkeywords={First keyword, Second keyword, More},
}

\begin{document}
\maketitle

\begin{abstract}
Protein sequence design is a challenging problem in protein engineering, which aims to discover novel proteins with useful biological functions. Directed evolution is a widely-used approach for protein sequence design, which mimics the evolution cycle in a laboratory environment and conducts an iterative protocol. However, the burden of laboratory experiments can be reduced by using machine learning approaches to build a surrogate model of the protein landscape and conducting in-silico population selection through model-based fitness prediction. In this paper, we propose a new method based on Batch Bayesian Optimization (Batch BO), a well-established optimization method, for protein sequence design. By incorporating Batch BO into the directed evolution process, our method is able to make more informed decisions about which sequences to select for artificial evolution, leading to improved performance and faster convergence. We evaluate our method on a suite of in-silico protein sequence design tasks and demonstrate substantial improvement over baseline algorithms.	
\end{abstract}


\section{Introduction}
\begin{figure}[h]
\vskip 0.2in
\begin{center}
\centerline{\includegraphics[scale=0.75]{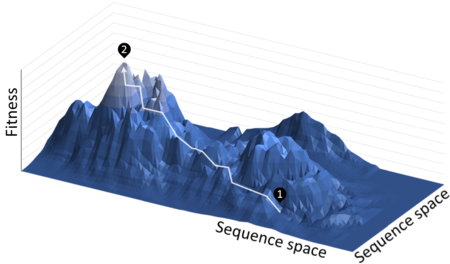}}
\caption{overview of landscape}
\label{landscape}
\end{center}
\vskip -0.2in
\end{figure}

Protein engineering is a rapidly growing field of biotechnology that aims to create or modify proteins to perform new or improved functions\cite{k2018protein}. The goal of protein engineering is to design proteins with specific desired characteristics, such as increased stability, specificity, and efficacy, or with new functions, such as the ability to bind to a target molecule by selecting sequences with high fitness scores\cite{turanli2012protein}. This can be accomplished through a variety of techniques, including rational design and directed evolution. However, the former approach necessitates a thorough understanding of the protein structure. As an alternative, directed evolution is inspired by natural selection pressure, which can be regarded as a searching task on fitness landscape\cite{romero2009exploring, yuan2005laboratory}. The advantage is that directed evolution can simulate natural evolution in a laboratory setting, through an iterative protocol\cite{packer2015methods}. During each iteration, a large number of variants are generated and evaluated through functional assays. Only the sequences with desired fitness scores are selected from the landscape and to form the next generation\cite{hopf2017mutation}.\\
\\
\noindent
The protein fitness landscape, shown in figure \ref{landscape}, is a crucial aspect of protein engineering, as it provides a mapping between protein sequences and their functional levels, or fitness scores. It represents the relationship between protein sequences and biological functions and is often represented as a high-dimensional surface\cite{hartman2019learning,ding2019deciphering}. The relationship between a protein sequence and its fitness can be influenced by many factors, including the protein's three-dimensional structure, interactions with other proteins and molecules, and the environment in which it operates. Therefore, understanding the protein fitness landscape is crucial in selecting protein sequences with improved fitness scores.
However, exploring this landscape can be a challenge as it requires expensive and time-consuming experiments or simulations\cite{thomas2021minding}. The traditional exploration mechanism used in directed evolution is a simple greedy strategy\cite{romero2009exploring}, starting from the wild-type sequence and accumulating mutations based on the fitness landscape. This unrestricted search can lead to sequences with high mutation counts that are far from the wild type, which can be challenging to synthesize and produce. Additionally, the training procedure of the surrogate model may have difficulty incorporating previous samples that are far from the current search region. Recent efforts have focused on using machine learning\cite{hie2022adaptive} to construct a surrogate model of the protein landscape and implementing model-guided search strategies. This approach effectively reduces the workload of laboratory experiments by performing in-silico population selection through model-based fitness predictions, but uncertainty and measurement errors are often neglected. To address this challenge,  we propose a novel approach that combines the strengths of Bayesian Optimization and Convolutional Neural Networks (CNNs) to model the protein fitness landscape and perform efficient sequence design. Our approach leverages the strengths of multiple CNNs to improve the scalability of Bayesian optimisation and accounts for the uncertainty in the relationship between protein sequences and their fitness, resulting a faster search for protein fitness values with improved accuracy.

\section{Related Work}
\textbf{Surrogate Model of Landscape.} The concept of the protein fitness landscape, which describes the relationship between a protein's sequence and its fitness score, has been used in protein engineering since 1932\cite{wright1932roles}, which provides a graphical representation of protein sequence and function. Exploring the entire protein fitness landscape can be difficult due to the time and cost involved, especially for large proteins with complex structure. Recently, machine learning approaches have shown promise in navigating protein fitness landscape\cite{freschlin2022machine,wittmann2021advances,narayanan2021machine}. It enables protein engineers more efficiently and effectively navigate these complex fitness landscapes and design proteins with desired functions\cite{luo2021ecnet,ogden2019comprehensive}. The learned model can be used to predict how mutations or modifications to the protein sequence will affect its fitness\cite{yang2019machine}. As the relationship between the sequence and its fitness can be influenced by various factors, Gaussian process is a popular choice\cite{romero2013navigating} , which helps account for the uncertainty in predicting the effects of mutations on protein function to guide the directed evolution of proteins with desired properties. Furthermore, the trained deep neural network can be used to screen large numbers of designed sequences in silico, without the need for wet-lab experiments\cite{ren2022proximal, shan2022deep, gruver2021effective}. \\
\\
\noindent
\textbf{Exploration Method.} Exploring sequence is another critical part in protein engineering\cite{gustafsson2001exploration}, that emphasized the need for efficient and systematic methods to explore sequence space, as the protein fitness landscape is vast and complex. The directed evolution has been widely utilised to search protein functions by evolving existing proteins \cite{jackel2008protein,dahiyat1996protein, bhattacherjee2009statistical}. Under such paradigm, different algorithms based on machine learning are adopted to guide evolutionary search and improve the sample efficiency\cite{wittmann2021advances,yang2019machine}. Dauparas et al.(2022)\cite{dauparas2022robust} proposed a neural network that can learn to predict the effects of mutations and explore new sequences that have desired properties. Brandes et al.(2022)\cite{brandes2022proteinbert} modified the architecture of language model BERT to develop the proteinBERT, which can generate new protein sequences with desired functional properties. Khan et al.(2022)\cite{khan2022antbo} proposed a combinatorial Bayesian optimisation to explore a large sequence space and optimize multiple properties simultaneously. The alternative approach of finding optimal sequences, such as model based reinforcement learning\cite{angermueller2020model}, which formulates biological sequence design as Markov decision problem. \\
\\
\noindent
\textbf{High Dimensional Optimisation.} Finding the optimal protein sequence is a difficult task due to the high dimensionality of the protein sequence space. Bayesian optimization can help search for new sequences by making informed decisions on which sequences to test based on their predicted properties\cite{belanger2019biological}. However, the high-dimensional biomedical datasets contain many features that are irrelevant to its biological function\cite{hu2016feature}. The utilization of large pre-trained models can aid in the reduction of the dimensionality of high dimensional optimization problems, particularly in the field of protein sequence design\cite{yang2022now}. This is achieved by providing a good level of predictive accuracy with a limited number of training examples. However, the use of large models in machine learning often requires significant computational resources, making the training process computationally expensive.Stanton et al.(2022)\cite{stanton2022accelerating} uses a denoising auto-encoder with BO to project high-dimensional data in a latent lower-dimensional space. The auto-encoder is trained to reconstruct the original sequence from a corrupted version, resulting in a compact representation of the sequences without relying on a pre-trained corpus. Our method utilizes a novel approach in which multiple simple Convolutional Neural Networks are employed instead of a single large model, which reduce the computational resource while still guaranteeing accuracy.

\section{Methodology}
\subsection{Problem Background}
\label{section:problem}
The problem of protein sequence design involves searching for a sequence, $s$, with specific properties within a high-dimensional sequence space, denoted by $V^{L}$. Here, $S$ represents the string of amino acids, and $L$ represents the desired length of the sequence. Let's define the protein fitness mapping function as $f$, which is a black box function that can be evaluated through laboratory experiments\cite{ren2022proximal}. The goal is to maximise $f: S^{L} \rightarrow R$ by modifying the starting sequence, $s_{0}$, such sequence should occur in nature. Specifically, we aim to find a mutant sequence $s^{*}$, such that it maximises the fitness score with the least amount of modification from the $s_0$.
\subsection{Bayesian Optimisation}
\begin{figure}[h]
\vskip 0.2in
\begin{center}
\centerline{\includegraphics[scale=0.5]{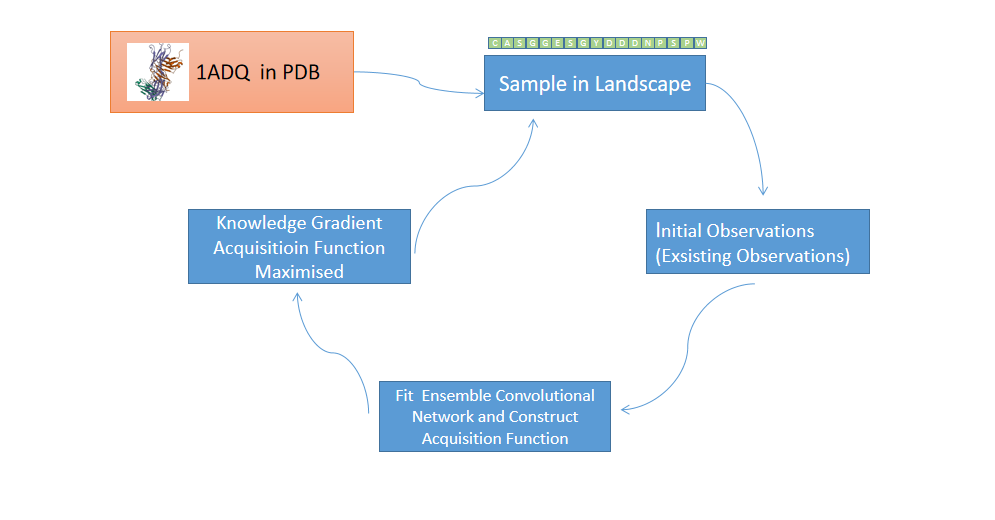}}
\caption{overview of landscape}
\label{pipeline}
\end{center}
\vskip -0.2in
\end{figure}

The optimization problem of finding a protein sequence with desired properties can be framed as a batch black-box optimization problem. Bayesian optimization has emerged as a promising solution for such expensive black-box optimization problems, as it is known for its data efficiency, allowing for the efficient exploration of the search space with limited measurements. The whole architecture of our method is shown in figure \ref{pipeline}.\\
\noindent
\\
Surrogate model and acquisition function are two main components of BO. Typically, gaussian process is a popular choice for building a surrogate model as it provides information about the uncertainty of the data through its mean and covariance. However, this approach can be computationally expensive for high dimensional data. Instead, we use an ensemble of 1D Convolutional Neural Networks as the surrogate model. The mean and variance of the results obtained from the ensemble CNNs serve as a substitute for the mean and covariance provided by the Gaussian process. This provides us with the ability to leverage the powerful representation learning capabilities of deep learning models to approximate the relationship between protein sequences and their fitness scores, while still maintaining the efficiency and scalability needed to handle high-dimensional optimization problems. The fitness model, $f_{\theta}$ is trained to predict fitness of mutant sequence, which minimises the expectation of the regression loss:
\begin{equation}
     L(\theta) = \mathbb{E}_{D} (f(s) - f_{\theta}(s))^2
\end{equation} where $D$ is the collection of observed sequences. The optimization process adjusts the parameters $\theta$ to minimize this loss, resulting in a surrogate model that can accurately predict the fitness of new sequences. By using this surrogate model, we can reduce the number of wet-lab experiments required, as we can first use the surrogate model to identify promising sequences before verifying their fitness through experiments.\\
\noindent
\\
Acquisition functions are used to guide the exploration of the black-box function in Bayesian optimization. They serve as a measure of the expected improvement of a candidate solution in comparison to the best observed solution thus far. The acquisition function operates by combining the mean and uncertainty estimates obtained from the surrogate model, and evaluating the potential improvement of candidate solutions. The acquisition function is then used to determine the next design point to sample, based on the criterion of maximizing the expected improvement. The choice of acquisition function can have a significant impact on the performance of Bayesian optimization, and different acquisition functions may be better suited for different types of optimization problems and objectives.The One-shot Knowledge Gradient (KSG), shown as \ref{eqn:kg}, 
\begin{equation}
\label{eqn:kg}
   \alpha_{KG}(s) = \mathbb{E}_{D}[\max_{s' \in S} \mathbb{E}[g(f(s')]] - \mu
\end{equation}
where $g \sim P(f(s') \vert D \cup D_{s})$ is posterior at $s'$, a new proposed sequence, $P$ is normally distributed in the default setting, $D$ is a collection of observed sequence and $D_{s}$  is a batch of sequences which are being observed and $\mu:= \max_{s} E[g(f(s)) | D]$, that is, the maximum expected value of the target function based on the current data set $D$. It is based on the idea of actively learning the best design point in a single-shot manner, as opposed to performing a full optimization over the entire search space\cite{wu2016parallel}. The KSG acquisition function evaluates the potential improvement of a design point by considering the uncertainty of the current model, as well as the expected improvement in the model's predictions after incorporating the results from that design point. This acquisition function has been demonstrated to be effective in various applications and has advantages over traditional acquisition functions such as Expected Improvement (EI) and Upper Confidence Bound (UCB). In this paper, we compare the performance of the One-shot Knowledge Gradient function with these two commonly used acquisition functions.

\subsection{Proximal Optimisation}
Drawing inspiration from the natural evolutionary process, where a protein can significantly enhance its fitness through mutations of a limited number of amino acids within a longer sequence, we restrict the search space to the vicinity of the wild type and aim to discover high-fitness mutants with limited mutation counts. Inspired by the concept of directed evolution, our approach prioritizes the selection of low-order mutants, which are sequences with a minimal number of mutations. To achieve this, we formulate a regularized objective function, $f_{\lambda}(s)$, that balances the trade-off between maximizing the fitness score of the sequence $f(s)$ and minimizing its hamming distance to the starting sequence $d(s, s_{0})$. This allows us to guide the sequence selection step towards finding high-fitness mutants with minimal mutations\cite{ren2022proximal, parikh2014proximal}. Therefore, we define new regularized objective function as follow:
\begin{equation}
    f_{\lambda}(s) = f(s) - \lambda \cdot d(s, s_{0})
\end{equation}
where $\lambda \ge 0$ is the regularization coefficient. Then we need to find a sequence $s^{*}$ such that $s^{*} := \argmax f_{\lambda}(s)$. $s^{*}$ can be seen as a function of $\lambda$. The choice of the value of $\lambda$ depends on the problem and the desired balance between exploring new regions and exploiting known regions of the sequence space. A higher value of $\lambda$ means a stronger focus on the starting sequence and nearby regions, while a lower value of $\lambda$ means less focus on the starting sequence and more exploration of the sequence space. 
\subsection{Model-guided Exploration}
In classical evolutionary algorithms, only the best-performing sequences are selected and mutated, which can limit exploration and overlook the properties of the natural protein fitness landscape. Instead, our algorithm extends the search beyond just optimizing the fitness score and focuses on exploring the proximal frontier to find high-fitness mutant sequences with low mutation counts. This mechanism is considered a regularization of the search space for protein sequences, and it encourages exploration of the sequence space to discover new and potentially optimal solutions.\\
\noindent
The overall procedure of the method is stated in Algorithm \ref{alg:batchbo}.
\begin{algorithm}
   \caption{Batch BO}
   \label{alg:batchbo}
\begin{algorithmic}
\STATE {\bfseries Input:} sequence $s_{0}$, model $f_{\lambda}(\theta)$,  observed data $D$, batch number $M$
\STATE Initialise the model parameter $\theta$
    \IF {not $D$}
    \STATE Initialise $D \gets \phi$
    \ENDIF
    \WHILE {$\mid D\mid < M$}
    \STATE Generate sequence batch $\{s_{i}\}$ based on $f_{\lambda}(\theta)$
    \STATE $D_{i} \gets D_{i-1} \cup (\{s_{i},f_{\lambda}(\theta)\})$
    \STATE Update $f$ based on $D_{i}$
    \ENDWHILE
\STATE {\bfseries Return:} a batch of optimal sequences
   
\end{algorithmic}
\end{algorithm}

Following the problem formulate in \ref{section:problem}, let $T$ be the number of rounds of interaction with the laboratory. At each round $t$, we propose a query batch containing $M$ sequence candidates and measure their fitness scores through wet-lab experiments. The measured protein fitness data are used to constantly refine the fitness model $\hat{f}_\theta$, where $\theta$ denotes the model parameter. In the context of batch black-box optimization, the exploration algorithm plays a crucial role in generating a query batch given the experimental measurements collected in previous rounds. To achieve this, a model-guided exploration algorithm leverages a fitness model $f_{\theta}$ and a dataset of measured sequences, $D$, to inform the selection of sequences for evaluation. Acquisition function evaluates the potential improvement of candidate sequences by considering the current best mean and uncertainty of the fitness model $f_{\theta}$. Based on the scores generated by the acquisition function, the most promising sequences are then selected for the next round of experiments. This iterative process continues until the termination criteria are met, and the algorithm returns the sequence with the highest predicted fitness score as the solution to the protein sequence design problem. \\
 \noindent
 \\
 Our method differs from classical evolutionary algorithms in that it takes into account not only the predicted fitness score of the candidate sequence, but also its distance from the starting sequence (the wild-type sequence) through mutations. While classical evolutionary algorithms perform a greedy selection based solely on the predicted fitness score of the candidate sequence, our method incorporates both factors into the selection process.

 \subsection{Surrogate Model of Landscape}
 We employed an improved method for modeling the protein fitness landscape that overcomes the limitations of traditional neural network approaches. Our method, an ensemble of Convolutional Neural Networks (CNNs), is specifically designed to capture the complex and highly non-smooth relationship between protein sequences and their fitness scores.The ensemble CNN provides not only a prediction of the fitness score of a given protein sequence, but also an estimate of its uncertainty. Furthermore, the ensemble CNN offers improved scalability compared to the Gaussian Process used in traditional Bayesian optimization, making it a more suitable choice for large-scale protein sequence design problems. 
\section{Experiments}

In this section, we present experiment to evaluate the performance of our method.Our method outperforms the state-of-the-art Proximal Exploration (PEX) method, which uses a mutation factorization network (MuFacNet) as a surrogate model. By leveraging the strengths of Bayesian optimization, machine learning, and the KG acquisition function, our method offers a promising approach to protein sequence design. To simulate the ground-truth protein fitness landscape, we use \textbf{Absolut!}\cite{robert2021one} as a replacement for the wet-lab measurements. Our exploration protocol can only interact with the simulated landscape through batch queries, and the sequence design algorithm cannot obtain any extra information about the black-box oracle.

\subsection{Performance Comparison}
We evaluate the effectiveness of our proposed method for designing high-scoring proteins through exploration of the fitness landscape. To assess the performance of our approach, we compare it against other two baseline algorithms.
\begin{itemize}
    \item \textbf{Random Search} is a common baseline used for comparison with other search strategies, which selects a previously measured sequence at random and then mutates it. Unlike other data-driven approach, random search does not use the model to guide the search strategy, but only to score the sequences.
    \item \textbf{PEX}\cite{ren2022proximal} is the algorithm that is a model-guided approach for protein sequence design that aims to reduce the need for costly in-lab experiments. The algorithm leverages the property of protein fitness landscapes that a concise set of mutations on the wild-type sequence can enhance the desired function. 
\end{itemize}
\begin{figure}[H]
\vskip 0.2in
\begin{center}
\centerline{\includegraphics[scale=0.33]{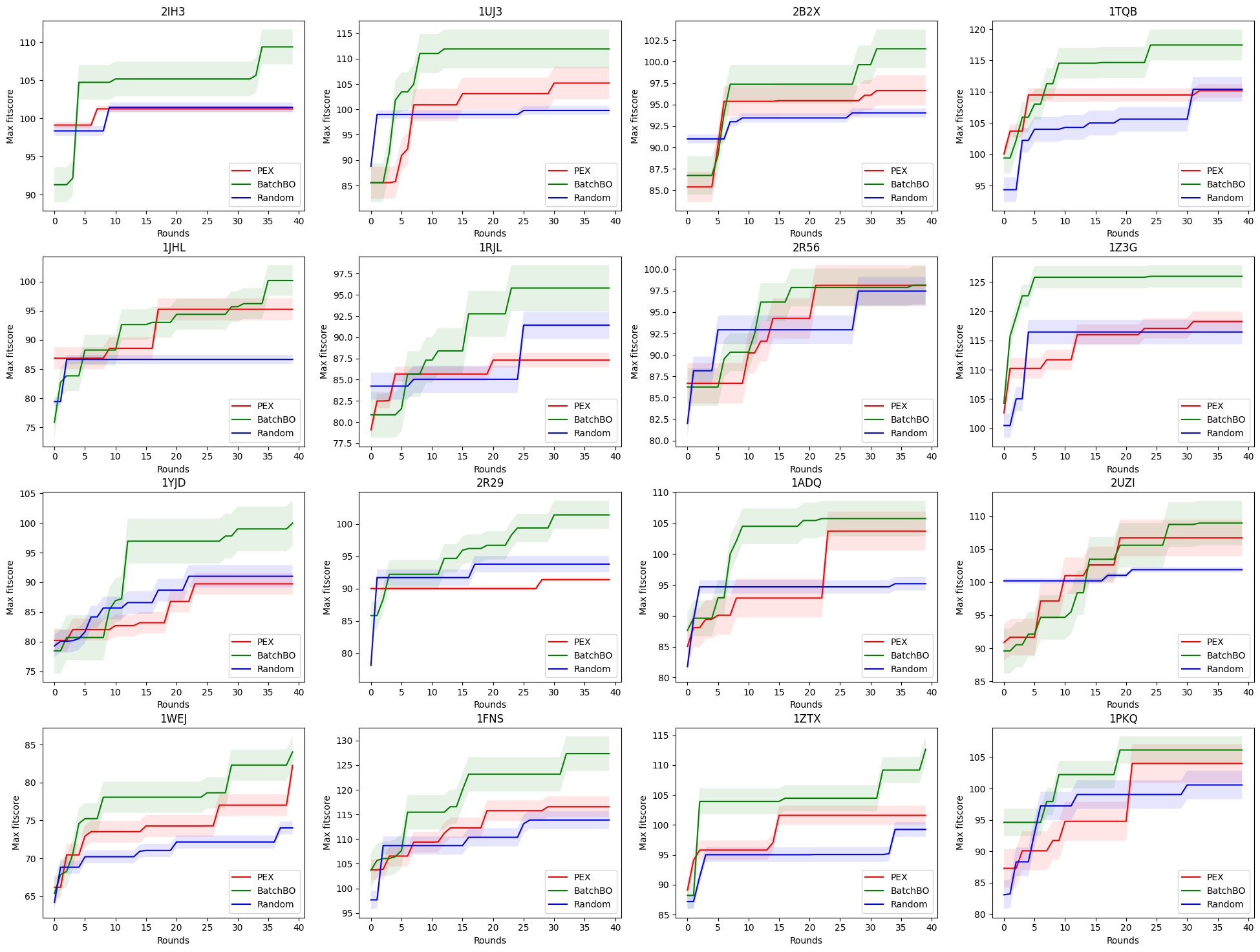}}
\caption{Learning curves on each PDB identifier, which generated by \textbf{Absolut!}. Each identifier contains 104,537 pairs of sequence and fitness energy.Each round of batch black-box query can measure the fitness
scores of 256 sequences.The evaluation metric used is the cumulative maximum fitness score among queried sequences, with all curves plotted from 40 runs using an ensemble of 1D convolutional neural network model architecture and KG as an acquisition function. The shaded region represents the standard deviation.}
\label{output}
\end{center}
\vskip -0.2in
\end{figure}
The result is shown in figure \ref{output}. Our experimental results demonstrate that our method outperforms the baselines in terms of achieving high fitness scores. Specifically, our method achieved the highest fitness score compared to other methods, after performing forty rounds of black-box queries, which improved the sample efficency of model-guided evolutionary search.
\subsection{Impact of Acquisition Function}

The effectiveness of an acquisition function depends on how well it balances exploration and exploitation. A good acquisition function should be able to guide the search towards promising regions of the fitness landscape while also exploring new regions. This is especially important in high-dimensional search spaces where the search can easily get stuck in local optima. In addition, the acquisition function should be computationally efficient to allow for fast evaluation of candidate solutions. 
To conduct an ablation study on acquisition function, I consider three types of acquisition functions: \textbf{UCB}, \textbf{EI} and \textbf{KG}. 
\begin{table}[H]
\label{sample-table}
\caption{Maximum Fitness Score Achieved by Each Method}
\begin{center}
\begin{tabular}{lcccr}
\toprule
Protein ID & 1V7M & 2DD8 & 2HFG \\
\midrule
CNN+UCB    & 113.15& 119.53 & 71.58 \\
CNN+KG & \textbf{113.48}&  \textbf{121.52}& \textbf{75.97} \\
CNN+EI    & 109.59&  120.97& 75.19 \\
RNN+UCB    & 106.67&  \textbf{115.45} &70.34       \\
RNN+KG     & \textbf{114.12}& 112.42& \textbf{75.97}\\
RNN+EI      & 101.99 & 110.82 & 75.19 \\
\end{tabular}
\end{center}
\end{table}
To enhance the comprehensiveness of our experiments, we also investigated the impact of incorporating RNN as a surrogate model along with different acquisition functions. We conducted forty rounds of searching and recorded the maximum score achieved during these rounds. The results of our experiment are presented in the Table \ref{sample-table}. Regardless of whether CNN or RNN was used, the method that employed the knowledge gradient function consistently outperformed the other methods in terms of achieving higher scores. This is because the knowledge gradient function selects a batch of points with the highest expected improvement values, which is particularly useful when searching for multiple high-performing solutions simultaneously. Consequently, it is a natural choice for batch optimization and explains why it achieved consistently better results in terms of maximum fitness score compared to other methods.

\section{Conclusion}
In conclusion, our proposed method for protein sequence design using Batch Bayesian Optimization has shown promising results in discovering novel proteins with useful biological functions. Our approach addresses the challenges of traditional directed evolution methods by incorporating a machine learning-based surrogate model and utilizing Batch BO to guide the search process. We also explored alternative models to the traditional Gaussian process surrogate model. We found that scalable models such as ensemble CNNs can offer a competitive alternative with the added benefit of faster computation times. Furthermore, we have shown that the choice of acquisition function plays a crucial role in the performance of the model-guided search. Our results suggest that the knowledge gradient function is particularly well-suited for Batch Bayesian Optimization, and can lead to improved sample efficiency and faster convergence. Overall, our work highlights the potential of combining machine learning with directed evolution for more efficient and effective protein sequence design.

\bibliographystyle{unsrt}
\bibliography{template}  






\end{document}